\documentclass[runningheads]{llncs}
\RequirePackage{silence}
\usepackage{graphicx}
\usepackage{comment}
\usepackage{amsmath,amssymb}
\usepackage{color}
\usepackage{url}

\usepackage{booktabs}
\usepackage{algorithm}
\usepackage{algpseudocode}
\usepackage{bm}
\usepackage{pifont}
\usepackage{colortbl}

\usepackage{hyperref}
\usepackage[capitalize]{cleveref}
\crefname{figure}{Fig.}{Figs.}
\Crefname{figure}{Figure}{Figures}
\crefname{table}{Tab.}{Tabs.}
\crefname{equation}{Eq.}{Eqs.}
\crefname{section}{Sec.}{Secs.}
\crefname{algorithm}{Alg.}{Algs.}

\usepackage{eccvabbrv}

\newcommand{\model}{MeSS}
\newcommand{\modelfull}{Mesh-based Scene Synthesis}
\newcommand{\bx}{\mathbf{x}}
\newcommand{\by}{\mathbf{y}}

\newcommand{\bc}{\mathbf{c}}

\newcommand{\bepsilon}{{\boldsymbol{\epsilon}}}

\newcommand{\bmu}{{\boldsymbol{\mu}}}

\newif\ifreview
\reviewfalse

\ifreview
	\usepackage{lineno}

	\linenumbers
\fi

\newcommand{\parhead}[1]{\par\noindent\textit{#1}\hspace{0.5em}\ignorespaces}

\begin{document}

%%%%%%%%%%%%%%%%%%%%% Add submission id, track, and title. %%%%%%%%%%%%%%%%%%%%%

\def\SubNumber{86}

% Track: Fast Review Track (resubmission of ECCV 2026 #7946)
\def\GCPRTrack{Fast Review Track}

\title{MeSS: City Mesh-Guided Outdoor Scene Generation with Cross-View Consistent Diffusion}

\ifreview
	% ANONYMOUS SUBMISSION FOR REVIEW
	\titlerunning{GCPR 2026 Submission \SubNumber{}. CONFIDENTIAL REVIEW COPY.}
	\authorrunning{GCPR 2026 Submission \SubNumber{}. CONFIDENTIAL REVIEW COPY.}
	\author{GCPR 2026 - \GCPRTrack{}}
	\institute{Paper ID \SubNumber}
\else
	% ===================== CAMERA READY AUTHOR BLOCK =====================
	% ORCIDs (encouraged): append \orcidID{xxxx-xxxx-xxxx-xxxx} after \inst{...}
	% for each author who has one.
	\titlerunning{MeSS: Mesh-Guided Outdoor Scene Generation}
	% Author list and order per the CMT record (13 authors).
	\author{%
	Xuyang Chen\inst{1,2}\thanks{Corresponding Author} \and
	Zhijun Zhai\inst{3} \and
	Kaixuan Zhou\inst{1} \and
	Zengmao Wang\inst{3} \and
	Jianan He\inst{1} \and
	Dong Wang\inst{1} \and
	Yanfeng Zhang\inst{1} \and
	Mingwei Sun\inst{1,3} \and
	Xuqin Wang\inst{1,2} \and
	R\"udiger Westermann\inst{2} \and
	Tao Wu\inst{1} \and
	Konrad Schindler\inst{4} \and
	Liqiu Meng\inst{2}%
	}
	\authorrunning{X. Chen et al.}
	\institute{Huawei Hilbert Research Center
	\and Technical University of Munich, Munich, Germany \\
	\email{xuyang.chen@tum.de}
	\and Wuhan University, Wuhan, China
	\and ETH Zurich, Zurich, Switzerland}
\fi

\maketitle              % typeset the header of the contribution

% Cover figure (above abstract)
\begin{figure}[t]
    \centering
    \includegraphics[width=\textwidth]{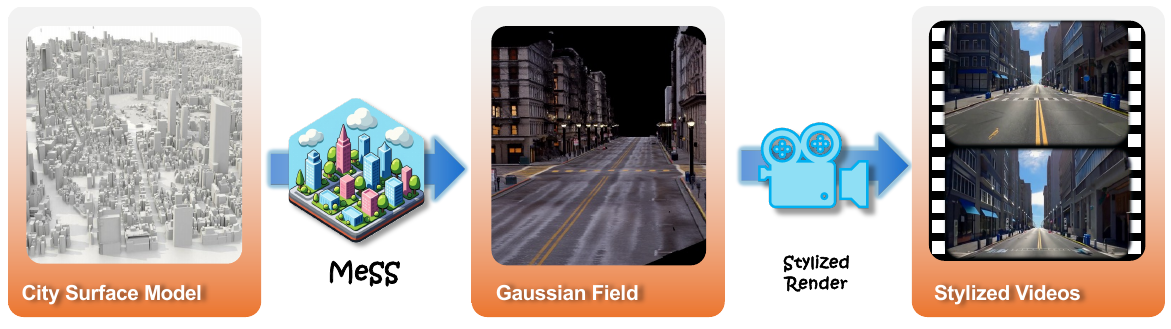}
    \caption{Starting from textureless urban meshes, our \model{} synthesizes high-quality Gaussian Splatting scenes with realistic appearance. After synthesis, these Gaussian scenes can be further rendered into stylized videos.}
    \label{fig:cover_figure}
\end{figure}

\begin{abstract}
Mesh models have become increasingly accessible for numerous cities; however, the lack of realistic textures restricts their application in virtual urban navigation and autonomous driving. Texturing them with generative models remains difficult: existing video-diffusion and outpainting-based scene generators drift in appearance over long camera trajectories and cannot guarantee pixel-level alignment with the given city geometry. This paper proposes \model{} (\modelfull{}), which converts a textureless city mesh into a high-quality, style-consistent 3D Gaussian Splatting (3DGS) scene. Our key insight is a control-distribution match: we train image ControlNets directly on mesh-rendered depth, normal, and semantic priors, so the control signals stay in-distribution for any deployed city mesh and yield tight geometry--appearance alignment. Building on this, three cross-view consistency mechanisms counter appearance drift: Cascaded Outpainting CtrlNets generate geometrically consistent sparse key views; Appearance Guided Inpainting (AGInpaint) densifies the scene with intermediate views; and Global Consistency Alignment (GCAlign) removes global inconsistencies such as exposure shifts. Concurrently with generation, the 3DGS scene is reconstructed by initializing Gaussian surfels on the mesh surface. On a city-scale benchmark from the UE5 City Sample Project, \model{} reduces cross-view LPIPS by 31\% and KID by 36\% relative to the best competing method on each metric, achieving an FID of 28.17, and it transfers zero-shot to a real LoD3 city mesh. Once synthesized, the scene can be rendered in diverse styles through relighting and style transfer.

\end{abstract}

% ---------------------------------------------------------------
% Main content

% ======== BEGIN sec/1_intro.tex ========
\section{Introduction}
\label{sec:intro}
Detailed city-scale mesh models are becoming widely available. Examples include 3DCityDB~\cite{3dcitydb}, Hamburger LOD3~\cite{hamburgLOD3_2025}, Ingolstadt LOD3~\cite{ingolstadt3d2020}, and 3DBAG~\cite{peters2022}, typically at Level of Detail 3 (LoD3) for buildings. However, these models often lack high-quality textures, which hinders downstream applications such as VR-based urban navigation, where street-level users are close enough to notice visual flaws.

Generative models offer a scalable and customizable way to enrich the visual appearance of such mesh models. Street-view capture with survey vehicles is prohibitively costly and prone to occlusions and illumination variability; generation is cost-efficient and lets users tailor scene aesthetics to their needs.

Existing generative scene methods, however, struggle to respect the geometry of a given city model. Recent works generate a city as NeRF~\cite{mildenhall2021nerf} or 3DGS~\cite{kerbl20233d} from city semantic \& height maps~\cite{chai2023persistent,lin2023infinicity,chen2023scenedreamer,xie2024citydreamer,xie2024gaussiancity,xie2025citydreamer4d}, but being supervised in bird's-eye view, their visual quality degrades significantly at street level. Other approaches~\cite{wang2024drivedreamer,li2024drivingdiffusion,zhao2024drivedreamer,gao2025vista} synthesize driving scenes with video diffusion models~\cite{blattmann2023stable}; focusing on foreground traffic participants and road infrastructure, they hallucinate the background---particularly buildings---without structural constraints, so the results often misalign with the actual city layout. StreetScapes~\cite{deng2024streetscapes} conditions on simplified building geometries and map layouts, and Cosmos-Transfer1~\cite{alhaija2025cosmos} injects sparse lidar-derived depth; yet their imprecise geometric inputs and feedforward conditional pipelines still fail to guarantee precise geometric alignment.

Approaches that operate directly on meshes or images come closer to our setting, yet still fall short at city scale. Texturing mesh models with 2D diffusion models~\cite{chen2023text2tex,metzer2023latent,chen2024scenetex,richardson2023texture} is limited to single objects or small scenes with low polygon counts. Perpetual view generation~\cite{yu2024wonderworld,yu2024wonderjourney,wang2024vistadream,chung2023luciddreamer} progressively outpaints RGB images and depth maps from a single image using 2D diffusion models~\cite{Rombach_2022_CVPR}, but gradually drifts away from the initial appearance and cannot generate consistent dense views.

In summary, one must solve two main challenges to generate city-scale scenes from mesh models:
\emph{(i)} Generate visual results aligning with given city surface models.
\emph{(ii)} Maintain a consistent appearance across long range without drifting.

Requirement \emph{(i)} motivates a \emph{control-distribution match}. Our image CtrlNets are trained directly on the depth, semantic, and normal priors rendered from city meshes. At deployment, the control signals from any city mesh---including real LoD3 models---therefore remain in-distribution and yield tight pixel-level alignment between geometry and appearance. Generic video-diffusion controllers, by contrast, are conditioned on depth estimated from real video and degrade on clean mesh-rendered control.

To meet requirement \emph{(ii)}, we organize generation as a two-staged \textit{warp-and-outpaint} framework with dedicated cross-view consistency mechanisms. In Stage I, \textbf{sparse key frames} are generated by two \textit{Cascaded Outpainting CtrlNets} that consecutively outpaint each frame from its predecessor, suppressing appearance drift over long ranges; the key frames are meanwhile spread as Gaussian surfels on the mesh surface to guarantee geometric alignment. In Stage II, intermediate views rendered from the Gaussian field---commonly exhibiting blurry artifacts or silhouettes---are repaired with \textit{Appearance Guided Inpainting} (AGInpaint). Finally, \textit{Global Consistency Alignment} (GCAlign) harmonizes remaining global discrepancies such as exposure shifts.

In summary, we make the following technical contributions:
\begin{itemize}
   \item
   We introduce \textbf{Cascaded Outpainting CtrlNets} conditioned on preceding frames to generate key frames that stay consistent over long camera trajectories.
   \item
   We devise an \textbf{Appearance Guided Inpainting} method to inpaint the occluded areas in the intermediate dense views, guided by surrounding known regions.
   \item On a city-scale benchmark rendered from the UE5 City Sample Project~\cite{epic2022citysample}, \model{} achieves state-of-the-art performance on 200-frame sequences spanning 200\,m camera paths, reducing cross-view LPIPS by 31\% and KID by 36\% relative to the best competing method on each metric with the lowest FID of 28.17, and transfers zero-shot to a real LoD3 city mesh (\cref{fig:lod3}).
\end{itemize}

% ======== END sec/1_intro.tex ========

% ======== BEGIN sec/2_related_work.tex ========
\section{Related Work}
\label{sec:related_work}

\subsection{Scene Generation from 2D Maps}
One line of work generates entire 3D cities from 2D semantic and height maps.
InfiniCity~\cite{lin2023infinicity} decomposes 3D scene generation into 2D map generation, map-to-voxel lifting, and voxel texturing via neural rendering.
SceneDreamer~\cite{chen2023scenedreamer} represents unbounded 3D landscapes via a BEV layout with height and semantic fields.
CityDreamer~\cite{xie2024citydreamer} and CityDreamer4D~\cite{xie2025citydreamer4d} refine this paradigm by decomposing rendering into specialized modules for distinct scene components, with CityDreamer4D further incorporating temporal consistency. Since these methods~\cite{xie2024gaussiancity,xie2024citydreamer,xie2025citydreamer4d} are supervised largely in bird's-eye view, they still struggle with layout stability and high-fidelity rendering at street level. \model{} instead conditions on an existing city mesh, so the generated scene conforms to the given city layout down to street-level detail.

\subsection{Perpetual View Generation}
Perpetual view generation~\cite{liu2021infinite} synthesizes views along arbitrary camera trajectories from a single image via a render-and-refine pipeline. This principle has been extended through GANs~\cite{li2022infinitenature} and diffusion models~\cite{cai2023diffdreamer,fridman2023scenescape}, with WonderWorld~\cite{yu2024wonderworld} enabling interactive 3D scene generation and VistaDream~\cite{wang2024vistadream} enforcing multiview geometry constraints during denoising. These approaches~\cite{chung2023luciddreamer,yu2024wonderjourney} all profit from monocular depth estimation~\cite{ranftl2020towards,bhat2023zoedepth,ke2024repurposing}, yet still exhibit rapid appearance drift in complex urban scenarios where outpainting error accumulates across occluded regions. Our Cascaded Outpainting CtrlNets adopt the same warp-and-outpaint principle but anchor every view to mesh-rendered controls, which suppresses this error accumulation.

\subsection{Scene Synthesis from Video Generation}
Video generators can synthesize approximately consistent views by conditioning on a single frame~\cite{yu2024viewcrafter,fan2024instantsplat,sun2024dimensionx} or interpolating multiple viewpoints~\cite{liu2024reconx,yu2024viewcrafter,sun2024dimensionx}. StreetScapes~\cite{deng2024streetscapes} addresses a task similar to ours with a two-frame AnimateDiff-based~\cite{guo2023animatediff} generator conditioned on rendered depth, height, and semantics, and Cosmos-Transfer1~\cite{alhaija2025cosmos} conditions video generation on sparse lidar-derived depth. However, training such models is expensive in data and compute, and their imprecise geometric control leads to temporal drift in long sequences. We instead build on image diffusion models whose controls are trained directly on mesh-rendered priors, keeping the control signals in-distribution at deployment without any video-model training.

\subsection{Test-time Guidance during Diffusion Sampling}
RePaint~\cite{lugmayr2022repaint} re-imposes known pixels at each reverse step to guide inpainting, but in Latent Diffusion Models~\cite{Rombach_2022_CVPR} the $8{\times}$ downsampling erases narrow mask boundaries. Recent works~\cite{yu2024wonderworld,viola2024marigolddc,shi2024dragdiffusion,koo2025flowdrag} instead back-propagate pixel-space losses through the latent at each denoising step, yet gradient-based guidance for RGB inpainting remains unexplored. Our AGInpaint fills this gap, inpainting intermediate views under the guidance of surrounding known regions.

% ======== END sec/2_related_work.tex ========

% ======== BEGIN sec/3_methods.tex ========
\section{Methods}
\newcounter{stagecounter}
\setcounter{stagecounter}{1}

\begin{figure}[t]
    \centering
    \includegraphics[width=\textwidth]{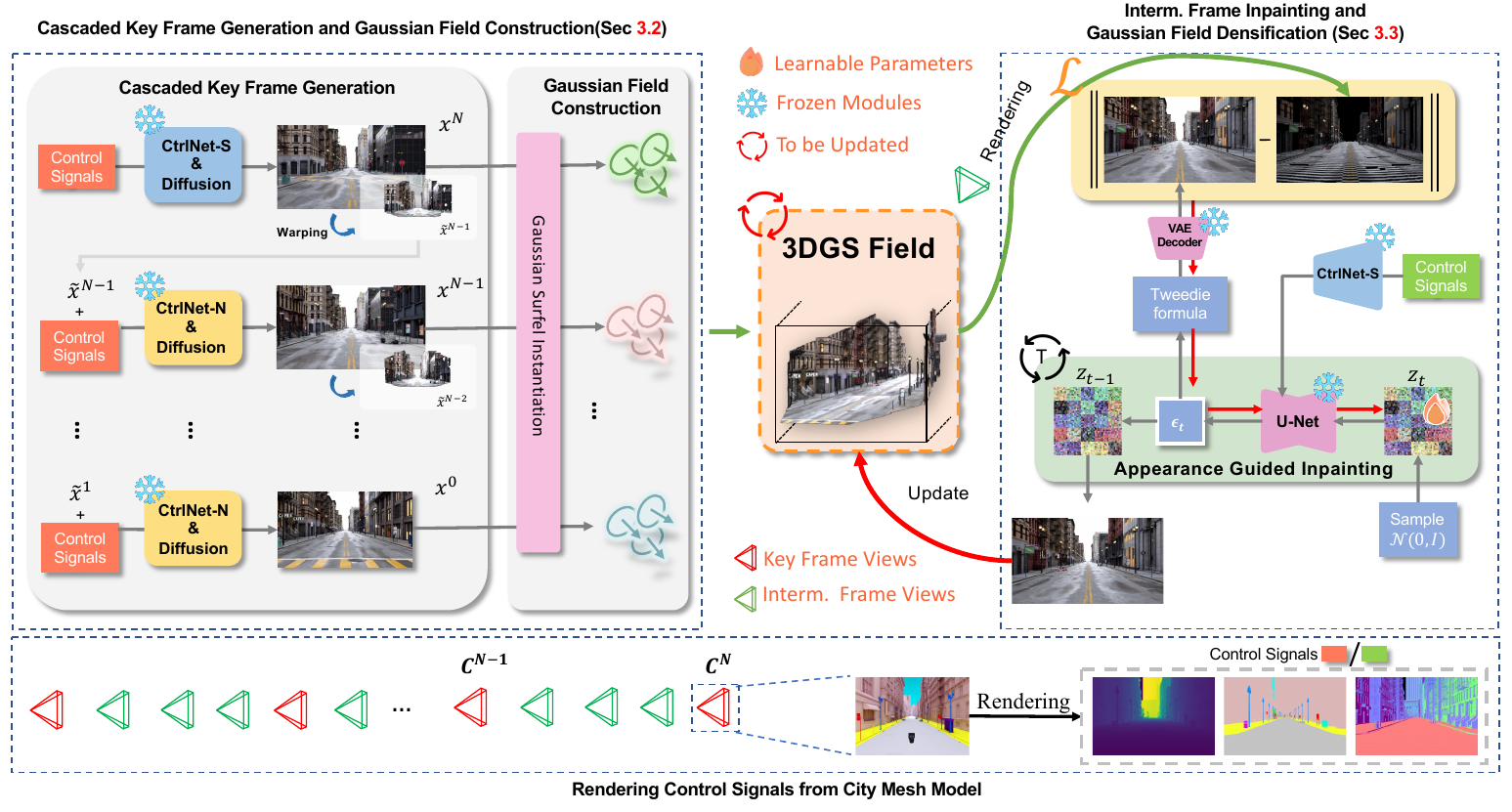}
    \caption{\textbf{Overview of the \model{} pipeline.} \textbf{Bottom:} given a city mesh and a camera path, we render control signals (depth, semantics, normals) at each viewpoint. \textbf{Left (Stage~I):} key views are generated autoregressively---starting from the last pose with \textit{CtrlNet-S}, then outpainting backwards via \textit{CtrlNet-N} with warped reference images---and instantiated as Gaussian surfels on the mesh surface to form a 3DGS field. \textbf{Right (Stage~II):} intermediate frames rendered from the Gaussian field are refined by Appearance Guided Inpainting, densifying the field with each inpainted view.}
    \label{fig:pipeline}
\end{figure}

The \model{} pipeline (\cref{fig:pipeline}) synthesizes views along a virtual camera path and reconstructs a Gaussian scene from them in a sparse-to-dense scheme; every generation step is conditioned on control signals rendered from the mesh, realizing the control-distribution match of \cref{sec:intro}. Given a 3D city map (\ie, a mesh model with semantic and instance labels but without texture) and a camera path of $M$ views, \textbf{Stage~I} generates $N$ sparse key frames by \textit{warp-and-outpaint} with two cascaded CtrlNets and instantiates them as Gaussian surfels on the mesh surface (\cref{subsec:Stage_I_Sparse_Scene_Construction}). \textbf{Stage~II} renders the intermediate views between consecutive key frames from this Gaussian field and repairs their silhouette artifacts with \textit{Appearance Guided Inpainting} (\cref{subsec:Stage_II_Dense_Scene_Construction}). Lastly, \textit{Global Consistency Alignment} (\cref{sec:Global_Consistency_Alignment}) harmonizes the appearance of surfels learned from different views.

\subsection{Preliminaries}

\parhead{Conditioned Image Diffusion Model.}
In a Latent Diffusion Model (LDM)~\cite{Rombach_2022_CVPR}, images are encoded by a pretrained autoencoder $(\mathcal{E},\mathcal{D})$ and a network $\hat{\bepsilon}_\theta$ is trained to denoise the latent; ControlNet~\cite{zhang2023adding} adds geometric control by training a copy conditioned on a signal $\bc$ (\eg, semantics or depth). In the forward process a clean latent is corrupted as $\mathbf{z}_t = \alpha_t\,\mathbf{z}_0 + \sigma_t\,\bepsilon$ with $\bepsilon\!\sim\!\mathcal{N}(\mathbf{0},\mathbf{I})$. Given the noise estimate, Tweedie's formula~\cite{efron2011tweedie} recovers the posterior mean of the clean latent,
\begin{equation}\label{eq_tweedie}
\hat{\mathbf{z}}_0 = \frac{\mathbf{z}_t - \sigma_t\,\hat{\bepsilon}_\theta(\mathbf{z}_t,\,\bc,\,t)}{\alpha_t},
\end{equation}
and $\hat{\by} = \mathcal{D}(\hat{\mathbf{z}}_0)$ gives a differentiable clean-image prediction at any step, which is essential for the gradient guidance in \cref{eq_guide_grad}.

\parhead{Gaussian Surfels from 2D Images.}
3D Gaussian Splatting (3DGS)~\cite{kerbl20233d} renders scenes via differentiable rasterization of Gaussian kernels; following~\cite{yu2024wonderworld,guedon2024sugar}, Gaussian \textit{surfels} flatten them into surface-aligned disks parameterized by position, orientation, tangential scales, opacity, and color. Given an RGB image with depth, one surfel is instantiated per pixel by back-projecting its position, copying the RGB color, aligning its orientation to the surface normal, and setting the scales proportional to $d/(\sqrt{2}\,f)$ ($d$ depth, $f$ focal length) for seamless coverage~\cite{yu2024wonderworld}; we give the full parameterization in the appendix.

\parhead{Multiview Consistency Sampling.}
VistaDream~\cite{wang2024vistadream} introduces training-free \textit{Multiview Consistency Sampling} (MCS): at each step $t$ it takes per-view clean estimates $\{\hat{\bmu}_t^{(n)}\}$, fits a temporary 3DGS, re-renders consistent images $\{\bar{\bmu}_t^{(n)}\}$, and rectifies the denoising direction by blending
\begin{equation}\label{eq_mcs}
\tilde{\bmu}_t^{(n)} = w_t\,\gamma_t^{(n)}\,\bar{\bmu}_t^{(n)} + (1-w_t)\,\hat{\bmu}_t^{(n)},
\end{equation}
where $w_t$ trades consistency against detail and $\gamma_t^{(n)}=\mathrm{std}(\hat{\bmu}_t^{(n)})/\mathrm{std}(\bar{\bmu}_t^{(n)})$ prevents exposure drift.

\parhead{Guided Diffusion Sampling.}
Sampling can be steered toward a partially observed target $\mathbf{y}_{\text{guide}}$ with visibility mask $\mathbf{M}_{\text{guide}}$~\cite{dhariwal2021diffusion,yu2024wonderworld,viola2024marigolddc}, by adjusting the noise estimate of the standard denoising step
\begin{equation}\label{eq_denoise}
\mathbf{z}_{t-1}=\mathrm{Denoise}(\mathbf{z}_t,\,t,\,\hat{\bepsilon}_t)
\end{equation}
with a guidance gradient:
\begin{align}
\hat{\bepsilon}_t &= \hat{\bepsilon}_\theta(\mathbf{z}_t,\,\bc,\,t) - s_t\,\mathbf{g}_t, \label{eq_guided_eps}\\
\mathbf{g}_t &= \nabla_{\mathbf{z}_t}\left\lVert (\hat{\mathbf{y}}_{t} - \mathbf{y}_{\text{guide}})\odot\mathbf{M}_{\text{guide}}\right\rVert^{2}, \label{eq_guide_grad}
\end{align}
where $s_t$ controls the strength and $\hat{\mathbf{y}}_{t}=\mathcal{D}(\hat{\mathbf{z}}_0)$ is decoded from the Tweedie estimate (\cref{eq_tweedie}).

\subsection{Stage \Roman{stagecounter}: Gaussian Field Construction from Key Views}\label{subsec:Stage_I_Sparse_Scene_Construction}
\parhead{Cascaded Key Frame Generation.}
Two multi-conditioned \textit{CtrlNets}~\cite{zhang2023adding} are introduced to generate $N$ key views in autoregressive fashion, moving backwards from the end to the start of the camera path\footnote{Views are indexed in driving direction.}. Both networks build on the standard ControlNet architecture but are extended to accept multiple control signals simultaneously; we illustrate their architecture in the appendix.
\textit{CtrlNet-S} receives three geometric conditions---depth $\boldsymbol{d}$, semantic segmentation $\boldsymbol{s}$, and surface normals $\boldsymbol{n}$---concatenated along the channel dimension and processed through convolutional layers before being fed into the ControlNet encoder. Since single-channel disparity maps may lose the ability to discriminate depth values in the far field, we encode depth as a colormap to preserve both near-field and far-field depth variations.
\textit{CtrlNet-N} extends this design with an additional reference image condition. The geometric maps follow the same processing path as in \textit{CtrlNet-S}, while the reference image is separately encoded by the frozen VAE encoder. The resulting latent features are element-wise added to the geometric condition features before being fed into the ControlNet, enabling the network to outpaint regions consistent with the reference view.

As illustrated in \cref{fig:pipeline} (left), starting from the last camera pose, \textit{CtrlNet-S} takes $\{\boldsymbol{d}^{N},\boldsymbol{s}^{N},\boldsymbol{n}^{N}\}$ at $\mathcal{C}^N$ to generate the initial view image $\bx^{N}$.
Having generated $\bx^{N}$, we warp the image content to its subsequent key view $\mathcal{C}^{N-1}$ via the known depth map. The reason for working backwards from the last frame is that, in a forward-facing camera, the warping will contract the pixel coordinates towards the image center. In this way, image generation artifacts due to limitations of the ControlNet, which mostly occur in the far field, will be diminished, whereas forward warping would amplify them and cause error build-up. To fill the peripheral regions where warped content is not available, \textit{CtrlNet-N} outpaints the missing values, taking the warped image $\tilde{\bx}^{N-1}$ as its reference condition alongside the geometric maps $\{\boldsymbol{d}^{N-1},\boldsymbol{s}^{N-1},\boldsymbol{n}^{N-1}\}$. The alternation between warping and outpainting is repeated until the first view of the sequence is reached. The next key frame is placed by thresholding the fraction of yet-unobserved (to-be-outpainted) area along the path, which corresponds to intervals of ${\sim}20$\,m on our trajectories.

However, the extended key frames generated with the described \textit{warp-and-outpaint} scheme may exhibit noticeable seams (\cref{fig:main}(b), left) due to different exposures. To remove them and achieve coherent appearance, we synchronize all key frames with a \textit{Global Consistency Alignment}, which will be discussed in \cref{sec:Global_Consistency_Alignment}.

\parhead{Gaussian Field Construction on the Mesh Surface.}
From the set of key frames $\{\bx^n\},n\in\left[1,N\right]$, a 3D Gaussian field is constructed. For the last frame $\bx^{N}$ (\ie the first generated image) we instantiate a Gaussian splat for every pixel. In subsequent key frames $\bx^{n}$, further Gaussians are added to fill in regions with low opacity values in the Gaussian field, in other words we also ``outpaint'' the Gaussians.

Since precise metric depth and surface normals are directly available from the mesh, we bypass any depth estimation and instantiate one surfel per pixel directly from the mesh depth and normals, without costly Gaussian field optimization. Additional details are provided in the appendix.

\setcounter{stagecounter}{2}
\subsection{Stage \Roman{stagecounter}: Intermediate Frame Inpainting and
Gaussian Field Densification}\label{subsec:Stage_II_Dense_Scene_Construction}

\parhead{Appearance Guided Inpainting.}
We render the constructed scene at intermediate novel views $\{\mathcal{C}^l\}$ between two key frames $\bx^n$ and $\bx^{n+1}, n\in[1,N)$.
Rendered from the coarse scene model, these novel views exhibit silhouettes---regions of low rendered opacity in the Gaussian field (see the appendix for an illustration). Novel views not included during the 3DGS reconstruction commonly suffer from holes and blurry textures~\cite{kerbl20233d}. We instantiate the guided diffusion sampling framework (\cref{eq_guided_eps,eq_guide_grad}) for RGB inpainting, yielding a training-free method termed \textit{Appearance Guided Inpainting} (AGInpaint).
Concretely, as shown in \cref{fig:pipeline} (right), the target signal $\mathbf{y}_{\text{guide}}$ is set to the rendered RGB image, and $\mathbf{M}_{\text{guide}}$ is obtained by thresholding the rendered opacity from the Gaussian field, thereby masking the silhouette regions that require inpainting. At each denoising step, the clean image $\hat{\mathbf{y}}_t = \mathcal{D}(\hat{\mathbf{z}}_0)$ is obtained via Tweedie's formula (\cref{eq_tweedie}), and the guidance gradient $\mathbf{g}_t$ (\cref{eq_guide_grad}) steers the noise estimate so that the prediction matches the known pixels outside the mask. This rectification is repeated $N_g$ times per step, incrementally updating $\hat{\bepsilon}_t$ with a learning rate $lr$, corresponding to gradient descent toward a local optimum that preserves sharp details and consistent textures.
The guidance is effective across the full sampling trajectory: at early timesteps it aligns the overall color palette with the reference, while at later timesteps the progressive refinement corrects fine-grained details.

The complete procedure, together with the adaptive rule that regulates the step size $s_t$ by scaling the gradient with the ratio of the mean absolute magnitudes of the current noise estimate and the gradient, is given in the appendix. All newly inpainted pixels are spread onto the mesh surface in the aforementioned way.

\subsection{Global Consistency Alignment}\label{sec:Global_Consistency_Alignment}
The Gaussian field produced by the above two stages still exhibits appearance artifacts such as brightness drift and outpainting mistakes. To resolve them, we design \textit{Global Consistency Alignment} (GCAlign), which applies the MCS framework (\cref{eq_mcs}) to our sparse-to-dense pipeline.

Concretely, we re-render the key views $\bx^{\prime\left(1:N\right)}$ from the current Gaussian field at viewpoints $\mathcal{C}^{\left(1:N\right)}$ and add $T_1$ steps of forward-diffusion noise ($T_1{=}9$ of our 50-step inference schedule) to obtain $\bx_T^{\left(1:N\right)}$. Tweedie's formula (\cref{eq_tweedie}) yields noise-free estimates $\tilde{\bx}_0^{\left(1:N\right)}$, which are blended with the rendered views following \cref{eq_mcs}: the rendered images $\bx^{\prime(n)}$ take the role of the 3DGS-consistent term $\bar{\bmu}_t^{(n)}$, and these noise-free estimates $\tilde{\bx}_0^{(n)}$ take the role of the diffusion prediction $\hat{\bmu}_t^{(n)}$, with the exposure-balancing factor $\gamma_t^{(n)}$ and trade-off weight $w_t$ defined identically.

The blended estimates serve to refine the Gaussian field; the updated scene is then re-rendered, and the process repeats across the denoising trajectory, progressively enforcing multi-view consistency while preserving fine details. Finally, the same procedure is applied to the sub-sequences of intermediate frames $\mathcal{C}^{\left(n:(n+1)k\right)}$ between consecutive key frames.

\begin{figure}[t]
    \centering
	    \begin{minipage}[b]{0.3\linewidth}
	        \centering
	        \includegraphics[width=\textwidth]{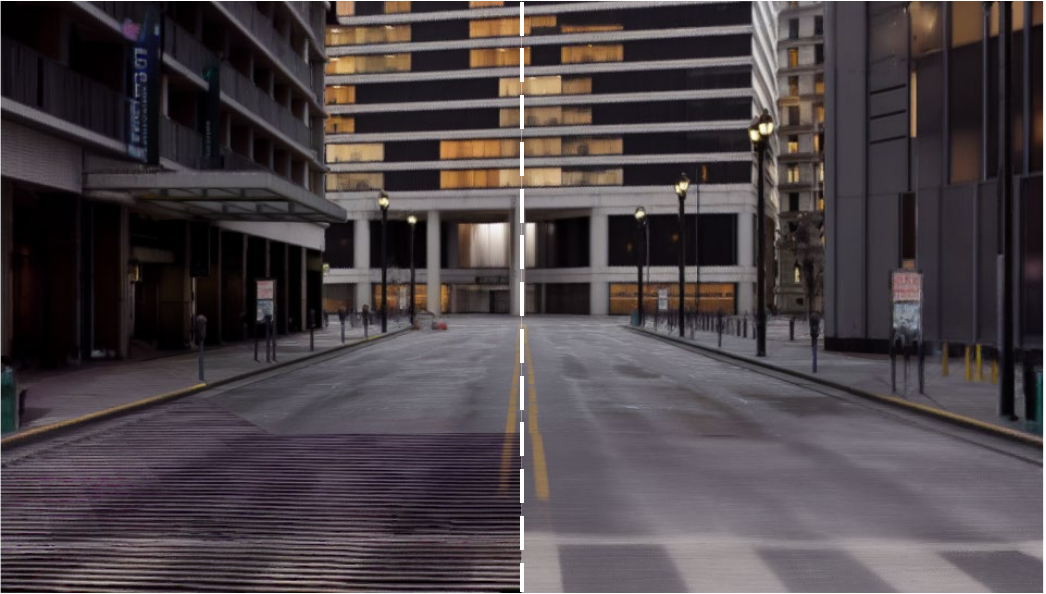}
	        \small (a)
	    \end{minipage}
	    \hspace{1em}
	    \begin{minipage}[b]{0.3\linewidth}
	        \centering
	        \includegraphics[width=\textwidth]{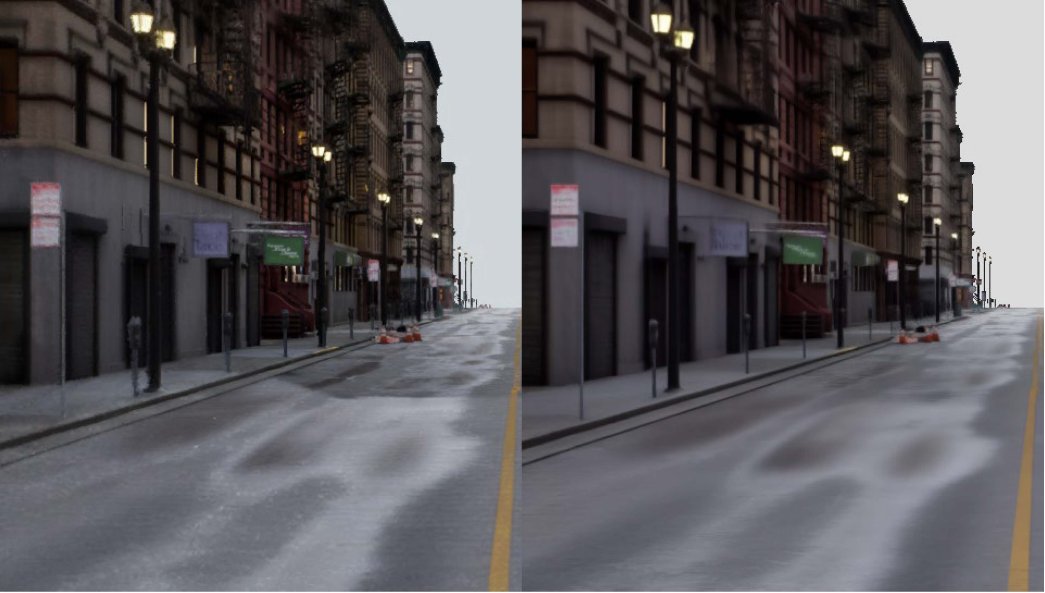}
	        \small (b)
	    \end{minipage}
    \caption{(a) Resample (left) vs.\ AGInpaint (right): AGInpaint inpaints slim regions more faithfully. (b) Without (left) and with (right) GCAlign: GCAlign harmonizes the seams caused by exposure differences.}
    \label{fig:main}
\end{figure}

% ======== END sec/3_methods.tex ========

% ======== BEGIN sec/4_experiments.tex ========
\section{Experiments}

\subsection{Data Preparation and Implementation}
\parhead{City Scene Data.}
We use the City Sample Project~\cite{epic2022citysample} in Unreal Engine~5~\cite{epic2021ue5}, which features two distinct cities with varied, realistic outdoor scenes. Camera sequences are rendered as RGB plus depth, normal, and semantic maps using modified tools from MatrixCity~\cite{li2023matrixcity} and EasySynth~\cite{ydrive2024easysynth}, at $960 \times 544$ with a $45^{\circ}$ field of view, simulating a forward-facing vehicle camera sampled at 1\,m intervals. In total, $16k$ images with paired control signals are obtained for training.

\parhead{Implementation Details.}
Both CtrlNets~\cite{zhang2023adding} are trained on frozen Stable Diffusion~1.5 (SD1.5)~\cite{Rombach_2022_CVPR} with a learning rate of $10^{-5}$ and a batch size of $128$ for $10k$ iterations. The reference image for \textit{CtrlNet-N} is randomly picked from $10\text{--}20$ meters ahead of the current view. For long-range generation, the sequence is divided into blocks of 200 frames processed autoregressively, each block starting from the last generated frame of the previous one.

To stabilize the one-step clean-signal estimation required by the guidance gradient (\cref{eq_tweedie}), we load LCM-LoRA weights~\cite{luo2023lcmlora} into SD1.5, whose self-consistency property~\cite{luo2023latent} provides a reliable noise-free prediction at any timestep. Guided rectification runs on the latent with an SDEdit-style~\cite{meng2021sdedit} $50\%$ noise initialization for $N_g=64$ iterations with a learning rate $lr=0.00375$ at each of 5 LCM denoising steps. Per-stage runtime and memory are reported in the appendix.

\subsection{Comparison with Other Methods}
We compare with methods from 3D city generation, video-based scene generation, and perpetual view generation. Generation quality is measured by FID and KID against ground-truth imagery, while cross-view consistency is evaluated via LPIPS on extended sequences generated from a real starting image. Because \model{} is, to our knowledge, the only outdoor mesh-based 3DGS generator, WonderWorld$^\dagger$ and VistaDream$^\dagger$ serve as directly matched baselines run end-to-end under identical geometric inputs, whereas StreetScapes and CityDreamer4D serve as paradigm reference points.

\parhead{Comparison with Perpetual View Generation.}
Both WonderWorld~\cite{yu2024wonderworld} and VistaDream~\cite{wang2024vistadream} rely on monocular depth estimation methods~\cite{bochkovskii2024depth,ke2024repurposing} to give depth value for placing Gaussians into 3D space, while our method is based on mesh geometry with precise metric depth. To form a fair comparison, we inject metric depth into their pipelines and align their inpainting diffusion model with ours; the altered versions are denoted WonderWorld$^\dagger$ and VistaDream$^\dagger$. For each camera path, we start the scene from the same initial view and outpaint it by moving the camera backwards for 200 meters with both baselines and our pipeline, producing 3200 images in total.

\begin{figure}[t]
	\centering
	\includegraphics[width=\textwidth]{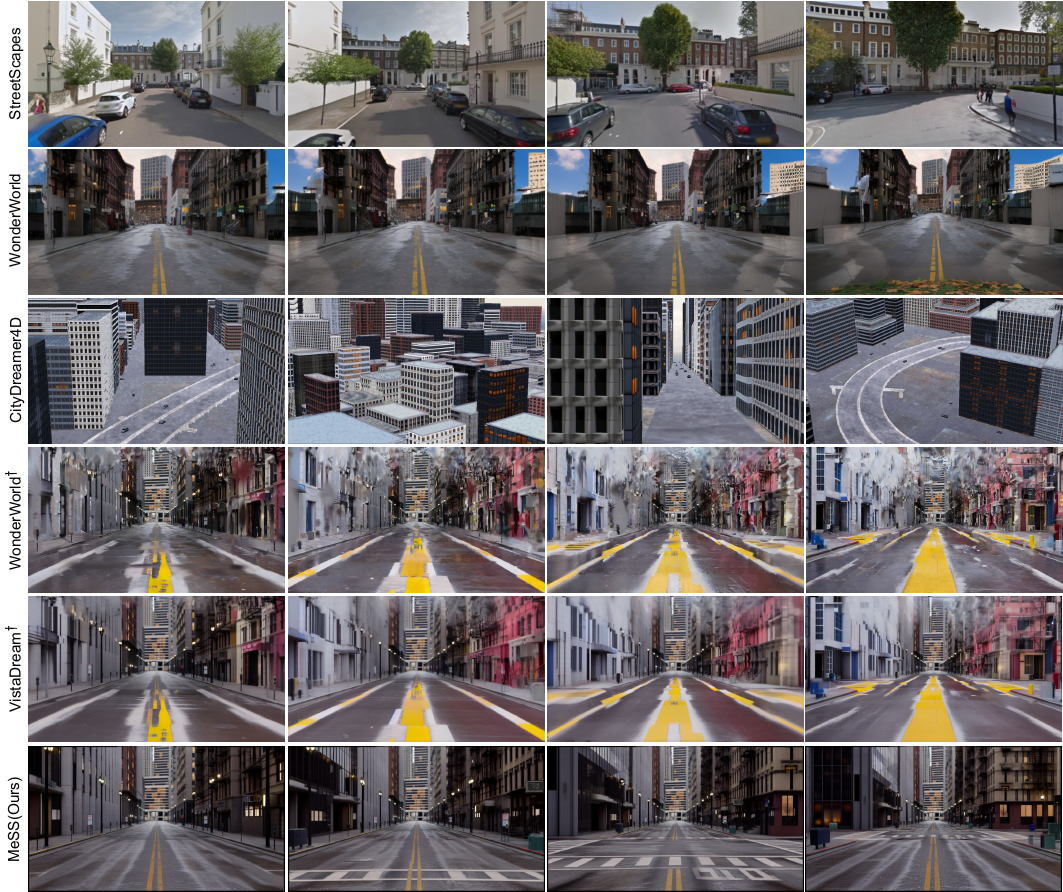}
	\caption{Qualitative comparison with scene-generation methods. Please zoom in to inspect details.}
	\label{fig:Comparison_with_Other_Methods}
\end{figure}

As depicted in~\cref{tab:quantitative_evaluations_perpetual_view}, our method achieves better metrics across the board. We visualize several intermediate frames in~\cref{fig:Comparison_with_Other_Methods}: for the same outpainting task, our results stay consistent and align well with the underlying geometry, whereas both baselines suffer from severe quality degradation caused by accumulated inpainting discrepancies. Notably, although VistaDream$^\dagger$ is equipped with Multiview Consistency Sampling, it shows no apparent improvement over WonderWorld$^\dagger$; consistency sampling pays off only when coupled with consistent inpainting, as in our pipeline. We also include the naive WonderWorld in~\cref{fig:Comparison_with_Other_Methods}: lacking any prior about the building geometry, it cannot extend facades and drifts to content irrelevant to the initial view, underscoring the difficulty of this task.

\begin{table}[t]
\centering
\small
\begin{minipage}[t]{0.52\linewidth}
	\centering
	\caption{\textit{Quantitative evaluations} on generated sequences.}
	\label{tab:quantitative_evaluations_perpetual_view}
	\resizebox{0.8\linewidth}{!}{%
	\begin{tabular}{lccc}
		\toprule
		& \textbf{LPIPS$\downarrow$} & \textbf{FID$\downarrow$} & \textbf{KID$\downarrow$} \\ \midrule
		CityDreamer4D~\cite{xie2025citydreamer4d}       & -              & 88.48          & 0.049\\
		WonderWorld$^\dagger$~\cite{yu2024wonderworld}   & 0.516          & 75.81          & 0.076\\
		VistaDream$^\dagger$~\cite{wang2024vistadream}   & 0.508          & 72.44          & 0.073\\
		StreetScapes~\cite{deng2024streetscapes}         & 0.519          & 29.93          & 0.025\\
		\rowcolor[gray]{0.9}
		\model{} (Ours)                                  & \textbf{0.348} & \textbf{28.17} & \textbf{0.016}\\
		\bottomrule
	\end{tabular}}
\end{minipage}
\hfill
\begin{minipage}[t]{0.45\linewidth}
	\centering
	\caption{\textit{Ablation studies}. Each row disables one pipeline component.}
	\label{tab:Ablation_studies}
	\resizebox{\linewidth}{!}{%
	\begin{tabular}{lccc}
		\toprule
		& \textbf{LPIPS$\downarrow$} & \textbf{FID$\downarrow$} & \textbf{KID$\downarrow$}\\
		\midrule
		w/o \textit{CtrlNet-N} & 0.382 & 54.54 & 0.053 \\
		w/o \textit{AGInpaint} & 0.422 & 51.12 & 0.046  \\
		w/o \textit{GCAlign}   & 0.346 & 26.25 & 0.013  \\
		\rowcolor[gray]{0.9}
		Full \model{}          & 0.348 & 28.17 & 0.016 \\
		\bottomrule
	\end{tabular}}
\end{minipage}
\end{table}

\parhead{Comparison with 3D City Generation.}
CityDreamer4D~\cite{xie2025citydreamer4d} is implemented on the Citytopia dataset~\cite{xie2025citydreamer4d}, which is also constructed based on the City Sample Project in Unreal Engine 5 (UE5), similar to ours. In comparison, our method demonstrates superior performance, as evidenced by lower FID and KID (\cref{tab:quantitative_evaluations_perpetual_view}). Furthermore, as illustrated in~\cref{fig:Comparison_with_Other_Methods}, their generated scenes exhibit notable limitations in visual quality upon closer inspection, particularly at street level. Specifically, the results lack critical urban elements such as streetlamps, traffic signs, and other street-level instances, significantly diminishing their realism and practical applicability.

\parhead{Comparison with Video-Based Generation.}
For StreetScapes~\cite{deng2024streetscapes}, we directly use the results reported in their paper since their code is not publicly available. However, forming a fair comparison is challenging, as their model is trained on a significantly larger dataset compared to ours\footnote{Even for their experiment on the London patch, they utilize approximately 20 times more data than we do.}. Given that our generated sequence is longer than theirs (200 frames vs.\ 64 frames), we evaluate their results within the range of 32 to 64 frames. In terms of quality, our results are on par with theirs, while we outperform them in terms of temporal consistency. This advantage can be attributed to our method's effective utilization of geometric priors, which plays a key role in achieving superior consistency.

\subsection{Real-World LoD3 Generalization}
\begin{figure}[htbp]
    \centering
    \includegraphics[width=0.9\linewidth]{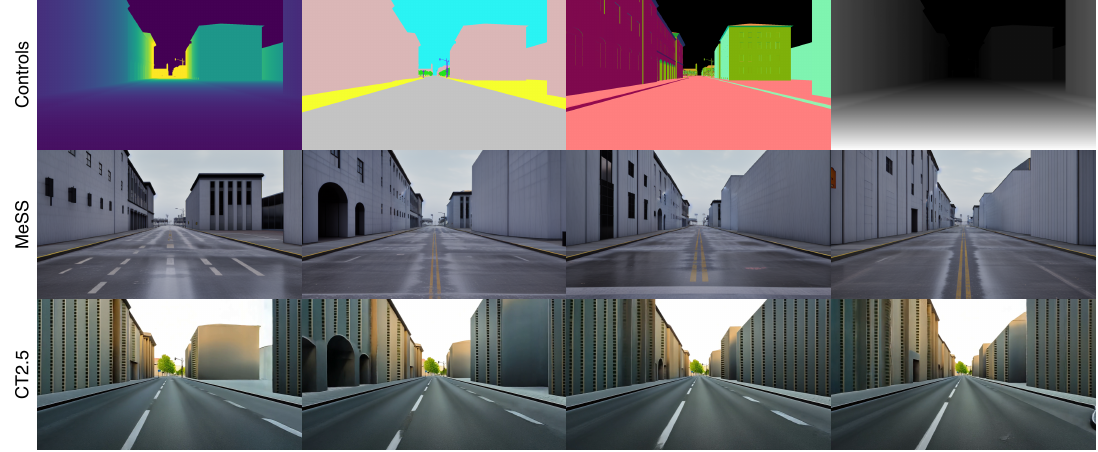}
    \caption{Zero-shot generalization to the real LoD3 mesh TUM2Twin~\cite{tum2twin}, no retraining. \textbf{Top:} mesh-rendered control signals---colormapped depth, semantics, normals, and the grayscale depth given to the video baseline. \textbf{Middle:} street-level views generated by \model{} along the camera path; facades follow the LoD3 geometry and remain consistent across views. \textbf{Bottom:} Cosmos-Transfer2.5 (CT2.5), a video-diffusion ControlNet conditioned on the same mesh geometry, preserves only a coarse layout with incorrect facade detail.}
    \label{fig:lod3}
\end{figure}

To test transfer beyond the synthetic training domain, we apply \model{} \emph{without any retraining} to TUM2Twin~\cite{tum2twin}, a public real LoD3 mesh of central Munich. Conditioned only on mesh-rendered depth, normal, and semantic priors, \model{} generates facades that follow the LoD3 geometry (\eg, doors and windows align with the rendered mesh) and stay consistent across views (\cref{fig:lod3}). This supports the control-distribution argument of \cref{sec:intro}: mesh-rendered priors are in-distribution for our CtrlNets. The residual failure mode is UE5-style texture drift on real facades, which a light Stage-I fine-tune on real imagery would close. The bottom row of \cref{fig:lod3} shows a head-to-head with the video-diffusion ControlNet Cosmos-Transfer2.5~\cite{nvidia2025cosmos25}, which preserves only a coarse layout; we analyze this comparison and give further details in the appendix.

\subsection{Ablation Study}
\parhead{Cascaded CtrlNets.}
For the outpainting of key views in Stage~I, we replace \textit{CtrlNet-N}, which is conditioned on the preceding frame, with \textit{CtrlNet-S}. \textit{AGInpaint} remains enabled. This setting differs from standard pipelines that combine warping and outpainting~\cite{yu2024wonderworld,yu2024wonderjourney}. As depicted in~\cref{tab:Ablation_studies}, the fidelity of generated frames degrades dramatically, and consistency also deteriorates because appearance drift accumulates during scene extrapolation.

\parhead{Appearance Guided Inpainting.}
We replace the \textit{AGInpaint} component in our method with a simpler approach called \textit{Resample}~\cite{deng2024streetscapes,lugmayr2022repaint}, which involves re-adding several rounds of noise at each denoising step to achieve a more homogeneous inpainted result. As demonstrated in \cref{tab:Ablation_studies}, the absence of \textit{AGInpaint} significantly impacts the generation quality. Through a visual comparison of the inpainted regions produced by \textit{AGInpaint} and \textit{Resample} (\cref{fig:main}(a)), we highlight the superior efficacy of \textit{AGInpaint}. Due to the downsampling of inpainting masks, \textit{Resample} struggles to fill in slim regions with Latent Diffusion Models, causing noticeable streak patterns.

\parhead{Global Consistency Alignment.}
As shown in~\cref{fig:main}(b), the generated scene exhibits noticeable lighting misalignment when \textit{GCAlign} is not applied, whereas integrating \textit{GCAlign} into our pipeline effectively harmonizes the seams. An interesting observation is that \textit{GCAlign} has a trade-off: while it improves visual coherence, it tends to introduce a slight blurring effect, leading to a loss of fine details. This is also reflected in the marginally better quantitative metrics when \textit{GCAlign} is disabled (\cref{tab:Ablation_studies}). Nevertheless, we consider this trade-off acceptable in favor of achieving a more visually consistent appearance, and it can be disabled when per-frame fidelity is preferred.

% ======== END sec/4_experiments.tex ========

% ======== BEGIN sec/5_conclusion.tex ========
\section{Conclusion}
We present \model{}, a pipeline for generating Gaussian fields from city mesh models, built on a \emph{control-distribution match}: CtrlNets trained directly on mesh-rendered priors yield tight alignment between geometry and generated appearance. By combining \textit{Cascaded Outpainting CtrlNets}, \textit{AGInpaint}, and \textit{GCAlign}, our method produces geometry-aligned scenes with coherent appearance across extended view sequences at low training cost, outperforming state-of-the-art scene generators on a city-scale benchmark and transferring zero-shot to a real LoD3 mesh.

\parhead{Limitations.} Our method is trained on high-detail meshes from the UE5 City Sample Project, and we show it transfers zero-shot to a real LoD3 mesh (\cref{fig:lod3}), preserving mesh-aligned layout while inheriting a UE5-style texture prior. Closing this residual texture-domain gap on a specific real city would benefit from a light fine-tune on real street imagery, which is far cheaper than the full CtrlNet training. In future work, integrating stronger base models (\eg, FLUX.1, SD3) and incorporating geometric priors into video diffusion are promising directions.

% ======== END sec/5_conclusion.tex ========

% ---------------------------------------------------------------
% Bibliography

\clearpage
\appendix
\section*{Appendix}

\noindent This appendix provides additional evidence and details for the main text. \cref{sec:supp_compute,sec:supp_consistency,sec:supp_lod3,sec:supp_ng,sec:supp_algorithm} consolidate the additional experiments and analyses; the remaining sections give architecture, texturing-baseline, and stylization details.

% Compute budget, cross-view metrics, LoD3 head-to-head, N_g early stop,
% and the AGInpaint algorithm.

\section{Compute Budget}
\label{sec:supp_compute}
\cref{tab:runtime} reports per-stage wall-clock and peak VRAM on a single RTX PRO 6000 for a 200-frame ($200$\,m) trajectory. Both stages are autoregressive, so cost scales roughly linearly with trajectory length and remains dominated by Stage~II; peak memory is unchanged for longer paths since sub-sequences are processed independently. Training each ControlNet takes ${\approx}10$\,h on eight RTX~A6000 GPUs.

\begin{table}[htbp]
\centering\small
\caption{Per-stage wall-clock and peak VRAM on a single RTX PRO 6000 for a 200-frame / $200$\,m trajectory.}
\label{tab:runtime}
\begin{tabular}{lrr}
\toprule
\textbf{Stage} & \textbf{Wall time} & \textbf{Peak VRAM}\\
\midrule
Stage~I    & 3\,min\,00\,s            & 5.6\,GB\\
Stage~II   & ${\approx}$17\,min\,10\,s  & ${\approx}$24.7\,GB\\
\midrule
\textbf{End-to-end} & ${\approx}$20\,min\,10\,s & ${\approx}$24.7\,GB\\
\bottomrule
\end{tabular}
\end{table}

\section{Cross-View Consistency and Novel-View Synthesis}
\label{sec:supp_consistency}
Cross-view LPIPS alone does not verify whether the \emph{same} surface keeps a consistent appearance across overlapping viewpoints. We therefore evaluate 12 shared trajectories against VistaDream$^\dagger$, the stronger matched baseline (\cref{tab:consistency}), using three additional measures. MEt3R~\cite{met3r} quantifies multi-view consistency directly on the rendered surface; KPM counts LoFTR keypoint inliers retained after RANSAC across cross-view offsets; and held-out novel-view PSNR, SSIM, and LPIPS are measured against UE5 ground truth. \model{} improves on all five metrics. We report held-out novel-view synthesis in place of FVD because \model{} emits a re-renderable 3DGS field rather than a video, so geometry-consistent novel-view rendering---not video temporal statistics---is the operative test of consistency.

\begin{table}[htbp]
\centering\small
\caption{Cross-view consistency and held-out novel-view synthesis against the matched baseline VistaDream$^\dagger$ (12 shared trajectories).}
\label{tab:consistency}
\begin{tabular}{lccccc}
\toprule
 & MEt3R$\downarrow$ & KPM$\uparrow$ & PSNR$\uparrow$ & SSIM$\uparrow$ & LPIPS$\downarrow$\\
\midrule
VistaDream$^\dagger$ & 0.155 & 1310 & 12.49 & 0.327 & 0.568\\
\rowcolor[gray]{0.9}
\textbf{\model{} (Ours)} & \textbf{0.129} & \textbf{1382} & \textbf{13.92} & \textbf{0.495} & \textbf{0.371}\\
\bottomrule
\end{tabular}
\end{table}

\section{Real-World LoD3: Comparison with a Video-Diffusion ControlNet}
\label{sec:supp_lod3}
The main text shows that \model{} transfers zero-shot to the real LoD3 mesh TUM2Twin~\cite{tum2twin}. Here we contrast it against Cosmos-Transfer2.5~\cite{nvidia2025cosmos25}, a strong video-diffusion ControlNet, conditioned on its own depth prior. \model{} generates facades that follow the LoD3 geometry and stay consistent across views, whereas Cosmos-Transfer2.5 preserves only a coarse layout with visibly wrong facade detail. This is consistent with a \emph{control-distribution mismatch}: Cosmos's depth ControlNet is trained on estimated depth from real video and is out-of-distribution on clean mesh-rendered depth. It does not by itself prove that retraining Cosmos's ControlNet on mesh-rendered priors could not close the gap---an experiment beyond the scope of this paper---so we position \model{} and video diffusion as complementary rather than as substitutes, and we state the corresponding claim in the abstract and Sec.~1 of the main text in terms of this control-distribution rationale rather than as a categorical image-vs-video position.

\section{AGInpaint Efficiency: $N_g$ Early Stopping}
\label{sec:supp_ng}
We also probed an NMSE-thresholded early stop on the inner gradient-guidance loop of \cref{alg:Apperance_Guided_Sampling}, in the spirit of exact-inversion samplers~\cite{hong2024exactInversion}. With a fixed threshold $\tau{=}10^{-3}$ across LCM denoising steps, the stop fires inconsistently---never at early steps (the residual stays above $\tau$ for all iterations) and immediately at late steps (halving the guidance budget)---so a single threshold does not yield a clean, monotone speedup. Searching nearby, we found instead that an SDEdit-style~\cite{meng2021sdedit} $50\%$ noise initialization with a reduced $N_g{=}64$ at the same 5 LCM steps matches the FID/KID of the original $N_g{=}100$ setting at ${\approx}30\%$ lower wall-clock. We therefore adopt this setting as the default; \emph{all numbers in the main text are reported with it}.

\section{Appearance Guided Sampling Algorithm}
\label{sec:supp_algorithm}
\begin{figure}[htbp]
    \centering
    \includegraphics[width=0.5\linewidth]{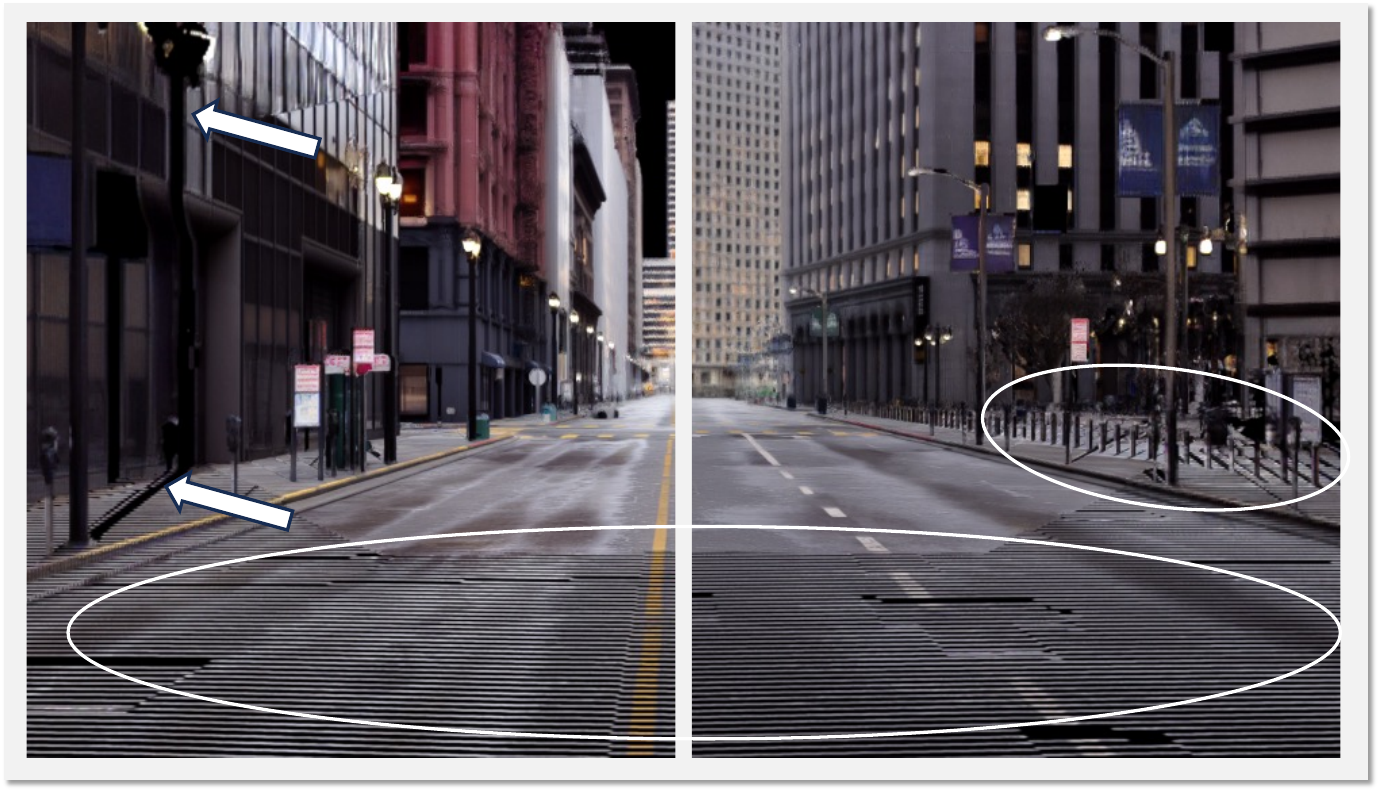}
    \caption{Silhouettes on novel views. The ellipses highlight the silhouette regions---areas of low rendered opacity in the Gaussian field---that AGInpaint is tasked to fill.}
    \label{fig:holes_of_intermediate_frames}
\end{figure}

As noted in the main text, intermediate novel views rendered from the coarse Gaussian field exhibit \emph{silhouettes}---regions of low rendered opacity that AGInpaint is tasked to fill. \cref{fig:holes_of_intermediate_frames} illustrates these regions on two example novel views.

\cref{alg:Apperance_Guided_Sampling} gives the complete AGInpaint procedure described in the main text. To adaptively regulate the step size $s_t$, we scale the gradient proportionally to the ratio of the mean absolute magnitudes of the current noise estimate and the gradient, accelerating convergence.

\begin{algorithm}[htbp]
  \caption{Appearance Guided Sampling} \label{alg:Apperance_Guided_Sampling}
  \footnotesize
  \begin{algorithmic}[1]
    \Require Target signal $\mathbf{y}_{\text{guide}}$, mask $\mathbf{M}_{\text{guide}}$, condition $\bc$, steps $N_g$, learning rate $lr$
    \State $\mathbf{z}_T \sim \mathcal{N}(\mathbf{0}, \mathbf{I})$
    \For{$t=T, \dotsc, 1$}
        \State $\hat{\bepsilon}_{t}^{(0)} \gets \hat{\bepsilon}_{\theta}\!\left(\mathbf{z}_t,\, \bc,\, t\right)$
        \For{$k=0,\dotsc, N_{g}-1$}
          \State $\hat{\mathbf{y}}_t \gets \mathcal{D}\!\left(\frac{\mathbf{z}_t - \sigma_t\,\hat{\bepsilon}_{t}^{(k)}}{\alpha_t}\right)$ \Comment{Tweedie estimate}
          \State $\mathbf{g}_t \gets \nabla_{\mathbf{z}_t}\left\lVert \hat{\mathbf{y}}_t\odot \mathbf{M}_{\text{guide}} - \mathbf{y}_{\text{guide}}\odot \mathbf{M}_{\text{guide}}\right\rVert^{2}$
          \State $s_t \gets lr \times \frac{\mathrm{mean}(|\hat{\bepsilon}_{t}^{(k)}|)}{\mathrm{mean}(|\mathbf{g}_t|)} $
          \State $\hat{\bepsilon}_{t}^{(k+1)} \gets \hat{\bepsilon}_{t}^{(k)} - s_t\,\mathbf{g}_t$
        \EndFor
        \State $\mathbf{z}_{t-1} \gets \mathrm{Denoise}(\mathbf{z}_t,\,t,\,\hat{\bepsilon}_{t}^{(N_g)})$
    \EndFor
    \State \textbf{return} $\mathcal{D}(\mathbf{z}_0)$
  \end{algorithmic}
\end{algorithm}

% Original supplementary sections carried over from the submission.

% --- Architecture, texturing-baseline, and stylization details ---
\section{Architecture of the Cascaded CtrlNets}
\label{sec:Implementation_details_of_CtrlNet-N}
\begin{figure}[t]
	\centering
	\includegraphics[width=0.8\linewidth]{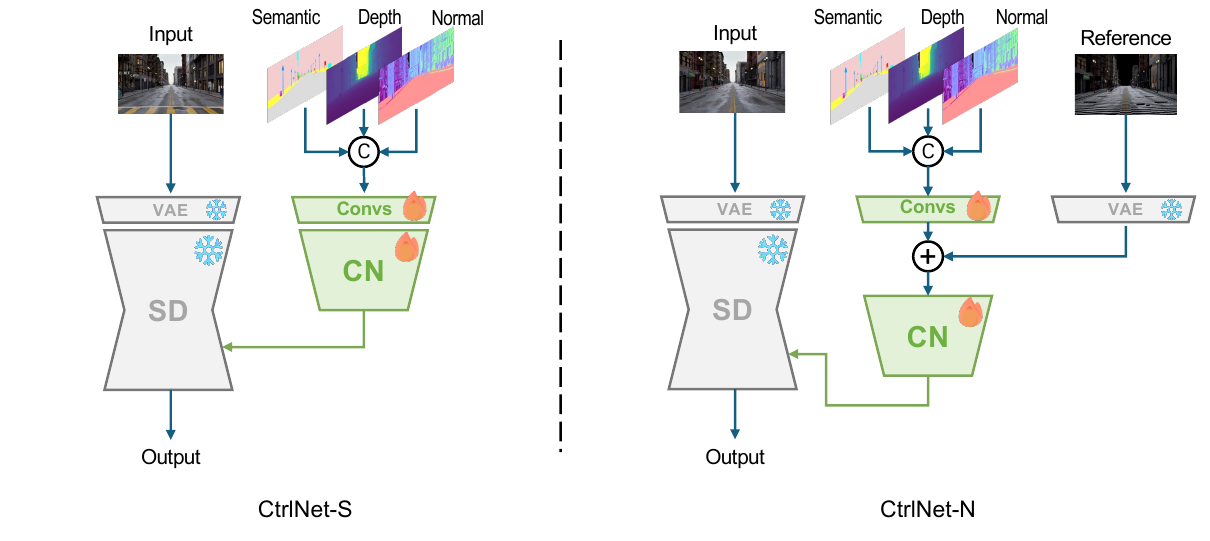}
	\caption{Architecture of our two cascaded CtrlNets. \textbf{Left:} \textit{CtrlNet-S} takes three geometric conditions (semantic, depth, normal) concatenated along the channel dimension. \textbf{Right:} \textit{CtrlNet-N} additionally takes a reference image encoded by the frozen VAE; the latent features are element-wise added to the geometric condition features before injection into the ControlNet. Snowflake and flame icons denote frozen and trainable components, respectively.}
	\label{fig:controlnet_arch}
\end{figure}

The architecture of our two cascaded CtrlNets is illustrated in \cref{fig:controlnet_arch}. Both build on the standard ControlNet~\cite{zhang2023adding} but are extended to accept multiple control signals simultaneously. The key difference of \textit{CtrlNet-N} from \textit{CtrlNet-S} lies in the processing of the reference image. Unlike the geometric control signals (depth, semantic and normal map), the reference image is first processed by the VAE encoder and then passed through several convolutional layers to align its feature dimension with the other control signal features before injection into the ControlNet.

\section{Trial on Mesh Texturing Method -- Text2Tex}
\label{sec:Setting_up_Text2Tex_for_outdoor_Scene_Generation}
\begin{figure}[t]
    \centering
    \includegraphics[width=\textwidth]{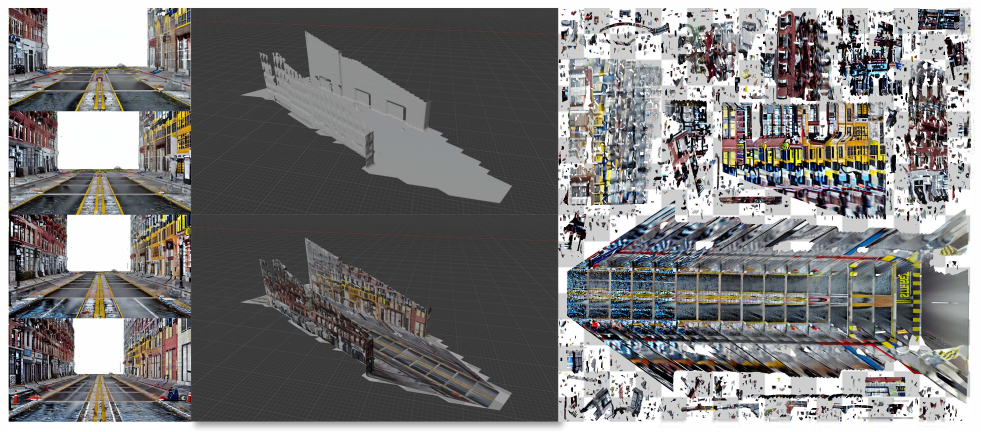}
    \caption{Text2Tex trial on a crossroad mesh from the City Sample Project.}
    \label{fig:Text2tex_result}
\end{figure}
Several authors have adopted diffusion-based generative models~\cite{ranftl2020towards,bhat2023zoedepth,ke2024repurposing,saharia2022image} for mesh texturing. TEXTure~\cite{richardson2023texture} and Text2Tex~\cite{chen2023text2tex} use depth-to-image diffusion models to texture meshes through inpainting, but can suffer from visible seams and gradual amplification of texture artifacts. Other methods, including Fantasia3D~\cite{chen2023fantasia3d}, Decorate3D~\cite{guo2023decorate3d}, SceneTex~\cite{chen2024scenetex}, and Latent Paint~\cite{metzer2023latent}, use Score Distillation Sampling (SDS)~\cite{poole2022dreamfusion,wang2023prolificdreamer}. Their costly test-time optimization limits them mainly to single objects or low-polygon indoor scenes, and they require training views that cover the relevant view field.

Text2Tex~\cite{chen2023text2tex} textures a mesh with a depth-to-image model by generating and inpainting the visible texels from predefined camera views. The final texture is represented as a UV-mapped RGB image. We adapt it to outdoor scene texturing, where City Sample Project instances are substantially more complex than Objaverse objects~\cite{deitke2023objaverse}. To make a 100-meter city block tractable, we apply Quadric Edge Collapse Decimation~\cite{garland1997surface} and retain only mesh faces visible from the camera views (the second column in \cref{fig:Text2tex_result}). We replace the original inpainting module with ours, extend the texture while moving the camera backwards, and increase the texture resolution from $3000 \times 3000$ to $4096 \times 4096$. Several randomly sampled views are then used to refine the texture. As shown in \cref{fig:Text2tex_result}, seams remain between successively inpainted regions, indicating limited scalability to large outdoor meshes.

\section{Distributing Flattened Gaussian Surfels On Mesh Surfaces}

Our approach for surfel generation draws conceptual inspiration from WonderWorld~\cite{yu2024wonderworld}, where depth and surface normals are predicted for novel views. However, since we directly utilize mesh geometry, we bypass this prediction step by retrieving accurate depth and normal information through rasterization.

Here, Gaussian surfels are parameterized with position $\mathbf{p}$, orientation in quaternion form $\mathbf{q}$, scales in orthogonal directions $\mathbf{s} = [s_x, s_y, \epsilon]$, opacity $o$ and RGB color $\mathbf{c}$. The Gaussian kernel at any spatial position $\mathbf{x}$ is given by:
\begin{equation}
	G(\mathbf{x}) = \exp\left(-\frac{1}{2} (\mathbf{x} - \mathbf{p})^T \mathbf{\Sigma}^{-1} (\mathbf{x} - \mathbf{p}) \right),
\end{equation}
where the covariance matrix $\mathbf{\Sigma}$ encodes the shape and orientation of the surfel and is defined as:
\begin{equation}
	\mathbf{\Sigma} = \mathbf{Q} \cdot \text{diag}(s_x^2, s_y^2, \epsilon^2) \cdot \mathbf{Q}^T,
\end{equation}
with $\mathbf{Q}$ derived from the quaternion $\mathbf{q}$. Our renderer uses the rasterization and alpha compositing process of 3D Gaussian Splatting (3DGS)~\cite{kerbl20233d}.

Given an image $\mathbf{I}$ of resolution $H \times W$, we construct $H \times W$ Gaussian surfels. For each surfel, the position $\mathbf{p}$ is directly derived from its corresponding 3D position on mesh, while it inherits color $\mathbf{c}$ from the pixel's RGB value. We assume that all surfaces are Lambertian, so $\mathbf{c}$ is treated as view-invariant. To avoid rendering artifacts such as undersampling or holes when zooming in, the scales along $x$ and $y$ axes are set as $d / {\sqrt{2} f_x}$, $d / {\sqrt{2} f_y}$, respectively. And $\epsilon$ is kept small enough but still larger than zero to avoid numerical errors. The opacity property is set to a constant 0.9. Similar to position, we align the surfel's normal with the mesh surface normal at the corresponding pixel. Then we receive the rotation matrix:
\begin{equation}
	\mathbf{Q}_z = \mathbf{n}, \quad \mathbf{Q}_x = \frac{\mathbf{u} \times \mathbf{n}}{\lVert \mathbf{u} \times \mathbf{n} \rVert}, \quad \mathbf{Q}_y = \frac{\mathbf{n} \times \mathbf{Q}_x}{\lVert \mathbf{n} \times \mathbf{Q}_x \rVert},
\end{equation}
with the global up-direction $\mathbf{u} = [0, 1, 0]^T$ used to ensure a consistent coordinate frame.

\section{Stylized Rendering}
\begin{figure}[t]
	\centering
	\includegraphics[width=\textwidth]{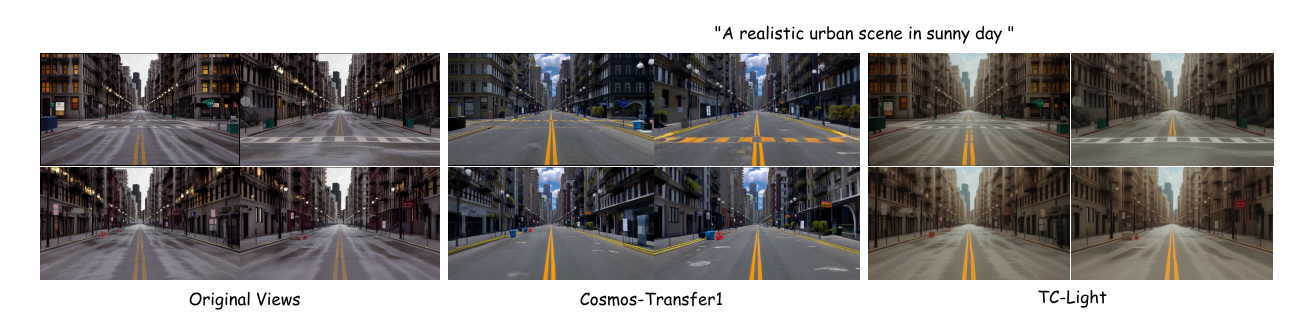}
	\caption{Stylized rendering results with Cosmos-Transfer1 and TC-Light. Please zoom in to check for details.}
	\label{fig:stylized_videos}
\end{figure}

The appearance of our generated scenes is strongly branded with the City Sample style, and simply tweaking the text prompt during view generation does not achieve stylization. We therefore customize the scene appearance through stylized rendering: given a camera path in the scene, we render a video and transfer it to a different style, either with the video relighting method TC-Light~\cite{liu2025tc} or via SDEdit~\cite{meng2021sdedit} with Cosmos-Transfer1~\cite{alhaija2025cosmos}. As depicted in \cref{fig:stylized_videos}, both complete this task by taking the original views as prior, with Cosmos-Transfer1 outperforming TC-Light in visual fidelity; the styled videos can be projected back to the Gaussian scene if desired.

The following prompts are leveraged for the inference of Cosmos-Transfer1 and TC-Light to achieve these results.

\begin{quote}
	\raggedright
	\textbf{Full prompt for Cosmos-Transfer1:}
	\par{\ttfamily A photorealistic scene of a quiet, empty urban street on a sunny day with some clouds in the sky. The wide road with yellow center lines stretches into the distance, flanked by tall buildings with classic architecture and street lamps. Some building fronts feature small potted shrubs, while a few windowsills display potted flowers or trailing leafy plants, adding touches of greenery and charm to the scene. The view is centered and symmetrical, creating a peaceful and cinematic atmosphere. Cool moonlight and warm streetlamp glows softly illuminate the buildings and pavement, casting gentle shadows. There are no people or vehicles, enhancing the stillness. Urban details like trash bins and cones add realism. The composition draws the eye toward a distant vanishing point.\par}
\end{quote}

\begin{quote}
	\raggedright
	\textbf{Full prompt for TC-Light:}
	\par{\ttfamily A photorealistic depiction of a calm, empty urban street at noon on a sunny day.\par}
\end{quote}

Further style variants (night, rain, snow) can be obtained with the following prompts:

\begin{quote}
	\raggedright
	\textbf{Cosmos-Transfer1 night time prompt:}
	\par{\ttfamily A photorealistic scene of a quiet, empty urban street at night under a clear sky with scattered clouds. The wide road with yellow center lines stretches into the distance, flanked by tall buildings with classic architecture and street lamps. Some building fronts feature small potted shrubs, adding a touch of greenery to the scene. The view is centered and symmetrical, creating a peaceful and cinematic atmosphere. Cool moonlight and warm streetlamp glows softly illuminate the buildings and pavement, casting gentle shadows. Urban details like trash bins and cones add realism. The composition draws the eye toward a distant vanishing point.\par}
\end{quote}

\begin{quote}
	\raggedright
	\textbf{TC-Light night time prompt:}
	\par{\ttfamily A photorealistic depiction of a calm, empty urban street at night under a moonlit sky.\par}
\end{quote}

\begin{quote}
	\raggedright
	\textbf{Cosmos-Transfer1 rainy day prompt:}
	\par{\ttfamily A photorealistic scene of a quiet, empty urban street during daytime under a cloudy, rainy sky. The wide road with yellow center lines stretches into the distance, flanked by tall buildings with classic architecture. Rows of street trees line both sides of the road, their wet trunks and leaves glistening slightly under the diffused daylight. The rain-slick asphalt reflects the buildings, trees, and urban elements, with scattered puddles creating mirror-like surfaces. Soft daylight filters through the overcast sky, producing muted shadows and emphasizing the textures of wet pavement and facades. Urban details like trash bins and traffic cones add realism. The composition is centered and symmetrical, guiding the eye toward a distant vanishing point softened by mist and rain.\par}
\end{quote}

\begin{quote}
	\raggedright
	\textbf{Cosmos-Transfer1 snowy day prompt:}
	\par{\ttfamily A photorealistic scene of a quiet, empty urban street during daytime in snowy weather. The wide road stretches into the distance, flanked by tall buildings with classic architecture and street lamps dusted with snow. Trees lining the sidewalks are covered with snow-laden branches, adding to the serene winter atmosphere. Some building fronts feature small potted shrubs, adding subtle touches of greenery amid the white landscape. The view is centered and symmetrical, creating a peaceful and cinematic atmosphere. Soft daylight reflects off the snow-covered pavement and buildings, casting diffuse shadows. Urban details like trash bins and cones partially covered in snow add realism. The composition draws the eye toward a distant vanishing point.\par}
\end{quote}

% ---------------------------------------------------------------
% One bibliography for the main text and appendices.
\bibliographystyle{splncs04}
\bibliography{main}

\end{document}